\documentclass[conference]{IEEEtran}
\IEEEoverridecommandlockouts
\usepackage{cite}
\usepackage{amsmath,amssymb,amsfonts}
\usepackage{algorithmic}
\usepackage{graphicx}
\usepackage{booktabs}
\usepackage{textcomp}
\usepackage{xcolor}
\usepackage{subfigure}
\usepackage{subcaption}
\usepackage{comment}

\def\BibTeX{{\rm B\kern-.05em{\sc i\kern-.025em b}\kern-.08em
    T\kern-.1667em\lower.7ex\hbox{E}\kern-.125emX}}

\usepackage{comment}
\usepackage{todonotes}

\begin{document}

\title{Event-based Neural Decoding for Neuroprosthetic Motor Control 
\thanks{Partially funded by the German Research Foundation (DFG, Deutsche Forschungsgemeinschaft) as part of Germany’s Excellence Strategy – EXC 2050/1 – Project ID 390696704 – Cluster of Excellence “Centre for Tactile Internet with Human-in-the-Loop” (CeTI) of TU Dresden, German Federal Ministry of Education and Research (BMBF), funding reference 16ME0729K, joint project "EVENTS" and the European Union within the programme Horizon Europe under grant agreement no. 101120727 (PRIMI).
}
}

\author{\IEEEauthorblockN{Khaleelulla Khan Nazeer\IEEEauthorrefmark{1}, Sirine Arfa\IEEEauthorrefmark{1}, Matthias Jobst\IEEEauthorrefmark{1}\IEEEauthorrefmark{2}, Richard George\IEEEauthorrefmark{1}, Christian Mayr\IEEEauthorrefmark{1}\IEEEauthorrefmark{2}}
\IEEEauthorblockA{\textit{\IEEEauthorrefmark{1}Chair of Highly-Parallel VLSI-Systems and Neuromorphic Circuits, Technische Universität Dresden, Germany} \\
\textit{\IEEEauthorrefmark{2}Centre for Tactile Internet with Human-in-the-Loop (CeTI), Dresden, Germany} \\
\{khaleelulla.khan, sirine.arfa, matthias.jobst2, richard\_miru.george, christian.mayr\}@tu-dresden.de
}
}

\maketitle

\begin{abstract}
A substantial number of patients experience diminished mobility due to disabilities, diseases, or accidents. Although modern prostheses, powered by deep neural networks, hold the promise of significantly enhancing the quality of life for these individuals, their widespread adoption is hindered by significant latency, energy consumption, and spatial requirements. Wired connections to external high-performance processors restrict patient mobility, while wireless connections limit the volume of information that can be transmitted to these processors. Spiking neural networks offer the potential for compressed communication and low-power inference, yet they often lag behind state-of-the-art deep learning models in various applications. In this study, we propose a high-performance neural decoding method that effectively balances task performance and efficiency. An event-based gated recurrent unit generates a sparse communication pattern with graded spikes, surpassing classical spiking neural networks in terms of task performance. Utilising an efficient training method and sparse inference, our model presents new opportunities for on-device neural decoding.
\end{abstract}


\section{Introduction}

The development of intra-cortical Brain-Machine Interfaces has accelerated in recent years, aiming to restore functional independence to individuals affected by paralysis and neurodegenerative diseases. To advance this field, the 2024 IEEE BioCAS Grand Challenge established a benchmark emphasizing neural decoders that combine high prediction accuracy with strict resource constraints inherent to implantable systems. Participants were evaluated using the Neurobench suite on non-human primate datasets and ranked by decoding accuracy and computational efficiency \cite{Yik2025Neurobench}.

Building on this foundation, the 2025 competition introduces an even more demanding scenario: closed-loop neural decoding. Here, algorithms must adapt and perform in real time, mirroring clinical neuro-prosthetic applications. The challenge consists of two tracks centered on a closed-loop center-out task where a virtual cursor must move from a central starting point to a randomly positioned target in a two-dimensional plane. An Online Prosthesis Simulator (OPS) generates neural activity from 96 directionally sensitive neurons in response to desired acceleration vectors, and decoders must infer velocity commands from this synthetic data. Performance is evaluated by time-to-target, the ability to maintain the cursor within the target area, and a low computational and memory footprint suitable for implantable devices.

Track 2 extends this framework by introducing controlled perturbations to emulate chronic degradation of neural recordings. Specifically, a random subset of probes is silenced to mimic signal loss due to increased impedance and scar tissue formation, while others are shifted to reflect activity from different neurons as probe positions drift.

To address these challenges, we leverage the Event-based Gated Recurrent Unit (EGRU)~\cite{subramoneyEGRU}, a neural network architecture designed to process sparse, event-driven data efficiently. Originally demonstrated on language modelling and gesture recognition tasks, EGRU naturally promotes activity sparsity, which can translate into substantial computational savings. Crucially, EGRU has been implemented on the SpiNNaker2 neuromorphic platform \cite{khan2024EGRUonS2}, where it demonstrated up to 18-fold improvements in energy efficiency over GPU-based inference. This combination of high predictive performance, low energy consumption, and suitability for real-time, sparse input streams makes EGRU a promising candidate for closed-loop neural decoding in implantable BMI systems.

The main contributions of this paper are:

\begin{enumerate}
    \item \textbf{Integration of EGRU into Reinforcement Learning:} First integration of EGRU into reinforcement learning for neural decoding, demonstrating that event-based recurrent architectures can effectively learn control policies in closed-loop settings.
    \item \textbf{On-Device-Efficient Decoder Design:} This approach achieves high task performance with only 2 EGRU units (2K parameters), demonstrating strong accuracy-efficiency trade-offs for implantable applications.
    \item \textbf{Robustness to Chronic Neural Signal Degradation:} Training and evaluating under realistic perturbations (signal dropout and tuning drift), showing resilience to electrode failure scenarios.
\end{enumerate}

\section{Methods}

Recurrent neural networks (RNNs) have long been used to give agents memory in sequential tasks~\cite{bakker2001RNN_RL}, especially under partial observability.
For example, incorporating an LSTM or GRU into deep reinforcement learning (DRL) agents allows the policy to integrate history and perform implicit state inference \cite{ni2022RNN_RL}.
In brain-computer interface (BCI) and neural decoding contexts, RNNs have likewise been effective, e.g. a closed-loop intra-cortical BMI with an RNN decoder outperforms classical Kalman filters \cite{sussillo2012RNN_RL}

Modern prosthetic decoding and robotics require low-power, event-driven computation.
Spiking neural networks (SNNs) mimic the brain’s sparse, asynchronous signals and are promising in resource-limited settings.
They can learn control policies with low-latency, asynchronous communication, but deep SNNs still lag behind dense DNNs in accuracy.
To address this, EGRU was introduced with GRU-level expressiveness that communicates sparsely like spiking neurons.
EGRU compute cost scales with activity: if only an $\alpha$ fraction of units fire, only $\alpha \times$ as many MACs are needed.

In the present work, we adopt the EGRU for neural decoding.
As in prior studies, we expect EGRU to provide a strong balance of performance and efficiency.
It can handle the temporal dynamics of neural signals (like other RNNs) but does so with far fewer active computations (like an SNN).
By exploiting EGRU’s activity sparsity and surrogate gradient training, our decoder can run on low-power hardware with significantly reduced effective MAC operations, making it well-suited for on-device prosthetic control.

To train our neural decoder effectively, we adopt a two-stage learning strategy: an open-loop pre-training phase followed by closed-loop reinforcement learning (RL).
In the pre-training stage, the model learns from synthetic trajectories generated offline followed by RL-based fine-tuning in a closed-loop environment.

\subsection{Model Architecture}
\label{section:EGRU-model}
\begin{figure}[tpb]
    \centering
    \includegraphics[width=0.8\linewidth]{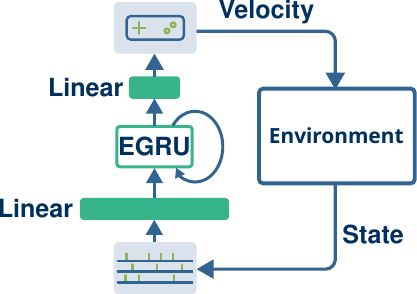}
    \caption{
    Model architecture consisting of EGRU and Linear projection layers.}
    \label{fig:model}
\end{figure}

The architecture consists of an initial linear layer that projects the input spikes to a 4-dimensional feature space, followed by an EGRU layer that processes the sequence data.
The model architecture is shown in Fig.~\ref{fig:model}.
EGRU consists of $n$ neurons with output $\mathbf{y}$ and state $\mathbf{h}$.
A sparse output $\mathbf{y} = (y_1, \dots, y_n)$ is generated from the GRU cell state $\mathbf{h} = (h_1, \dots, h_n)$ via a thresholding mechanism.
These "graded spikes" of the output can take any floating-point value above the threshold.
The output of the EGRU is then passed through a linear head to produce the final output.
$\mathbf{y}$ is the sparse output of the EGRU layer that's fed back into the model as recurrent input.
The hidden state $\mathbf{h}$ represents the internal state of each cell, which is not communicated between units.
We use $n = 2$ for benchmarking the results shown in Table~\ref{table:EGRU2-multiple-neurons}, resulting in a very low footprint of just 2K parameters.
The layer sizes can be summarised as 96-4-2-2, where 96 input channels feed into a 4-neuron linear layer. 
Then, a 2-unit EGRU is processed, and the output is transformed into a two-dimensional velocity at each timestep using a linear layer at the output.

\subsection{Reward Function}
\label{section:reward}
We designed the reward function to provide a smooth, differentiable signal for reaching a target in a continuous control task.
It computes the Euclidean distance between the agent’s current position and the target, then uses a sigmoid function to softly indicate whether the target has been reached (with a sharpness parameter controlling the transition).
The total reward is the sum of three components: (1) a completion reward that increases when the agent finishes close to the target, (2) a time bonus that rewards reaching the target quickly, and (3) a distance penalty that discourages being far from the target.
This design encourages both accuracy and efficiency in reaching the target while avoiding abrupt changes in the reward landscape.
We used back-propagation to update the model weights based on the reward.
A conventional RL algorithm was used, where the agent receives rewards at each time step, accumulating them as it progresses.
The hidden state is not reset between timesteps; instead, it is reset only at the beginning of the episode, when the environment is also reset. 

\subsection{Open loop pre-training}
In the pre-training phase, we initialise the model using a large dataset of randomly generated trajectories sampled from the environment.
This synthetic data is created by simulating random agent movements, ensuring broad coverage of the state and action spaces.
Pre-training on this data serves two key purposes: First, it exposes the model to diverse input patterns, effectively acting as an exploration phase before reinforcement learning.
Second, it enables the use of efficient CUDA kernels provided by the EGRU implementation, significantly accelerating the training process.
By leveraging this pre-training strategy, the model acquires a robust initialisation that facilitates faster convergence and improved performance during subsequent RL training on task-specific data.
We used trajectories of length equal to 4 time-steps, using longer trajectories did not yield any benefit.

\subsection{Closed-Loop Reinforcement Learning}
During the reinforcement learning (RL) phase, the EGRU model is further trained by interacting with the environment.
In each episode, the agent receives the current neural state and outputs a velocity command.
Actions are drawn from a Gaussian distribution centred on the model’s prediction, with standard deviation controlled by an exploration parameter, $\epsilon$. This promotes early exploration and gradually shifts to exploitation as $\epsilon$ decays.

The environment returns the next state, a shaped reward, and a termination signal.
The agent accumulates rewards and updates its parameters after each episode via gradient ascent on the mean reward (minimising the negative reward).
Performance is tracked using total reward, final distance to the target, and time taken.

To evaluate robustness under degraded recording conditions, we introduced controlled perturbations before evaluation.
A configurable fraction of synthetic neurons was either silenced or had their tuning properties reassigned by resampling preferred directions and firing rates.

These perturbations were also applied during training, with affected neurons sampled anew each epoch to prevent reliance on fixed patterns.
This encouraged generalisation to unseen disruptions. Table~\ref{table:T2-EGRU2-multiple-neurons} summarises performance under these conditions, showing the model maintained a high success rate despite increased variability.

To further assess resilience, we performed a sweep over increasing fractions of perturbed neurons—combining silencing and redistribution—to simulate chronic electrode failure and drift. Figure~\ref{fig:sweep-results-all-neurons} shows the impact of these combined challenges on decoding performance.

\section{Experimental Results}
To evaluate our EGRU-based neural decoder, we conducted a series of experiments designed to assess performance, robustness, and efficiency under varying conditions.
Our training pipeline follows a two-stage process: open-loop pre-training followed by closed-loop reinforcement learning.

\subsection{Training setup}

We start with an open-loop pre-training phase using 5,000 synthetic trajectories, each four time steps long, to broadly sample the state-action space. This phase runs for 30 epochs with supervised learning to initialise model weights before policy optimisation.

Next, models train in a closed-loop reinforcement learning setup for 600 episodes, using the shaped reward from Section~\ref{section:reward}.
Gaussian exploration noise with an initial epsilon of 0.5, decaying to 0, is applied to predicted velocity commands.

A grid search explores key hyperparameters such as EGRU $n$, learning rate, and epsilon to balance performance and efficiency.

\subsection{Evaluation}
After training, models are evaluated over 50 trials with randomised targets and initial states. 
Table~\ref{table:EGRU2-multiple-neurons} summarises the performance of three EGRU-based decoders on Track 1, on each of the three different neuron models. We demonstrate minimal footprint while reaching and maintaining the target on average within less than 1\,s, and with 100\% success within the episode.

In Track 2, models face controlled neural perturbations (channel silencing, distribution shifts) to assess robustness, with randomised perturbation parameters per trial. Results are summarized in Table~\ref{table:T2-EGRU2-multiple-neurons}. Similar to Track 1, all models reach and maintain the target with 100\% success rate, with only a slight increase of the time to target.

\begin{figure}
    \centering
    \includegraphics[height=0.22\textwidth, keepaspectratio]{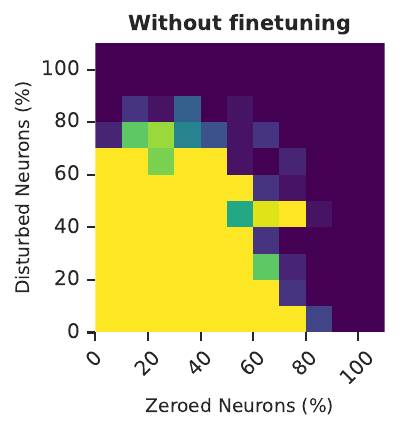}
    \includegraphics[height=0.22\textwidth, keepaspectratio]{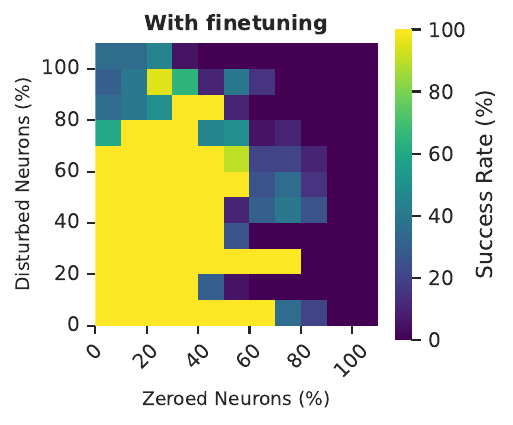}
    \caption{Heatmap of the success rate vs percentage of neurons with their distributions shifted or zeroed. EGRU is inherently robust against perturbations and can be further enhanced with fine-tuning.}
    \label{fig:success-heatmap}
\end{figure}

We also analyze the effect of the perturbations on the success rate. Figure~\ref{fig:success-heatmap} presents a heatmap of success rates as a function of the percentage of neurons affected by silencing or distributional shift.
The EGRU decoder maintains high performance even under substantial perturbations, demonstrating inherent robustness, which can be further improved through targeted fine-tuning on perturbed data.

\subsection{Comparison to LSTM-based decoder}
We compare our EGRU decoder against a standard LSTM-based decoder with comparable parameter counts in Table~\ref{tab:EGRU-LSTM-comparision}.
Both models are trained and evaluated under identical conditions.
While LSTM achieves similar accuracy and slightly faster time to target, it requires significantly more compute due to its dense activations and more complex inner structure and lacks the event-driven sparsity of EGRU.

\begin{table}[htp]
\centering
\caption{Performance of EGRU model as described in Section~\ref{section:EGRU-model} on multiple neuron models.}
\begin{tabular}{@{}lccc@{}}
\toprule
\textbf{Metric} & \textbf{Neuron 1} & \textbf{Neuron 2} & \textbf{Neuron 3} \\
\midrule
Footprint & 2172 & 2172 & 2172 \\
Connection Sparsity & 0.00 & 0.00 & 0.00 \\
Activation Sparsity & 0.001 & 0.041 & 0.002 \\
Effective MACs & 4126.04 & 3947.88 & 4351.32 \\
Effective ACs & 5581.92 & 6492.40 & 5030.80 \\
Dense Synaptic Ops & 38579.32 & 37426.92 & 40712.12 \\
Avg Time to Target (s) & 0.8972 & 0.8704 & 0.9468 \\
\bottomrule
\end{tabular}
\label{table:EGRU2-multiple-neurons}
\end{table}

\begin{table}[htp]
\centering
\caption{Performance of the EGRU model on multiple neuron models under Track 2 perturbations, with 50\% of neurons retuned to simulate distribution shifts and 40\% silenced to simulate signal loss.}
\begin{tabular}{@{}lccc@{}}
\toprule
\textbf{Metric} & \textbf{Neuron 1} & \textbf{Neuron 2} & \textbf{Neuron 3} \\
\midrule
Footprint & 2172 & 2172 & 2172 \\
Connection Sparsity & 0.00 & 0.00 & 0.00 \\
Activation Sparsity & 0.004 & 0.051 & 0.026 \\
Effective MACs & 4678.64 & 3780.76 & 5100.96 \\
Effective ACs & 5130.48 & 3783.12 & 5446.24 \\
Dense Synaptic Ops & 43782.32 & 35964.92 & 48099.52 \\
Avg Time to Target (s) & 1.02 & 0.83 & 1.12 \\
\bottomrule
\end{tabular}
\label{table:T2-EGRU2-multiple-neurons}
\end{table}

\begin{table*}[htp]
    \centering
    \caption{Comparison of EGRU and LSTM-based decoders. Both models have comparable footprints.}
    \begin{tabular}{@{}cccccccc@{}}
    \toprule
        Model Type & Hidden Dim & Activation Sparsity & Effective MACs & Effective ACs & Dense Ops & Avg Time to Target (s) \\
        \midrule
        EGRU & 2 & 0.09 & 4020 & 7441 & 38777 & 0.90  \\
        EGRU & 4 & 0.08 & 9612 & 7212 & 43408 & 0.87  \\
        \midrule
        LSTM & 2 & 0.00 & 4763 & 6765 & 36305 & 0.82  \\ 
        LSTM & 4 & 0.00 & 12081 & 7068 & 43431 & 0.82  \\
        \bottomrule
    \end{tabular}
    
    \label{tab:EGRU-LSTM-comparision}
\end{table*}

\begin{figure*}[htp]
    \centering
    \includegraphics[width=0.95\textwidth]{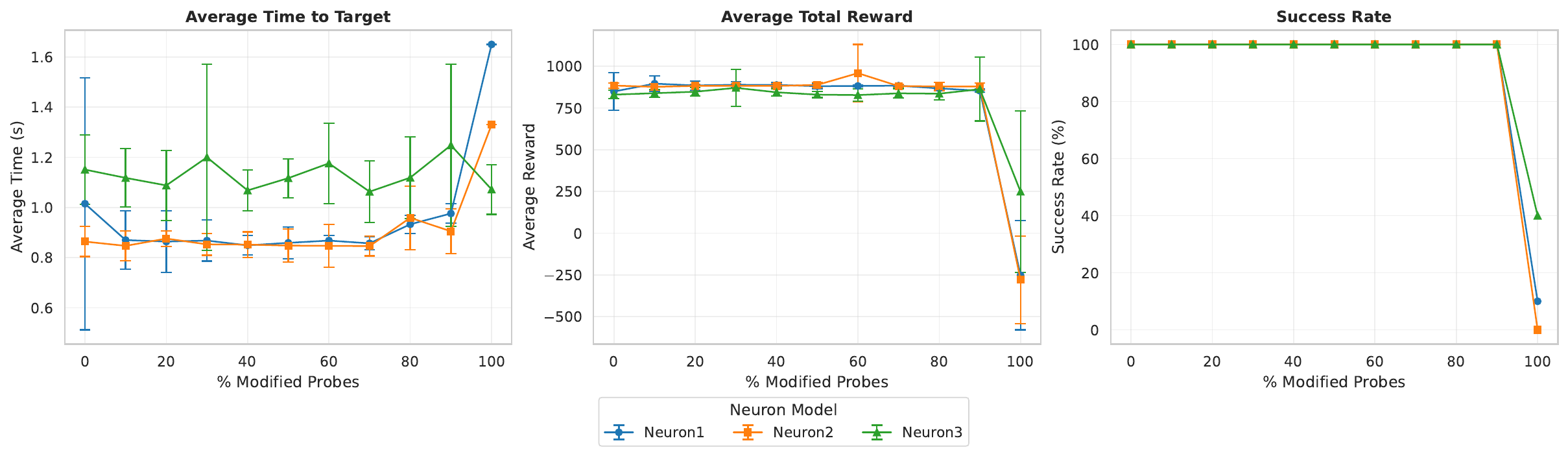}
    \caption{
        Performance of the EGRU ($n=2$) decoder across all neuron models under varying fractions of perturbed probes.
        Each curve represents results for a distinct neuron configuration, highlighting the effect of probe modifications on average time to target, total reward, and success rate.
    }
    \label{fig:sweep-results-all-neurons}
\end{figure*}

\section{Discussion}

In this work, we successfully trained the EGRU decoder to achieve robust closed-loop neural decoding across both benchmark tasks. In Track 1, the decoder attained an average time to target below 1 second across all neuron models, demonstrating efficient control performance and low latency. In Track 2, we systematically evaluated the model under progressively increasing fractions of perturbed probes, combining signal dropout and distributional shifts up to 100\% modification. Even under severe perturbation levels, the decoder maintained a high success rate and stable reward up to approximately 80\% probe modification, confirming its resilience to chronic signal degradation. 
A similar sized LSTM decoder reaches the target position only slightly faster, at the cost of a higher number of computations.
Detailed results are reported in Table~\ref{table:EGRU2-multiple-neurons} and Table~\ref{table:T2-EGRU2-multiple-neurons}, the performance trends are visualised in Figure~\ref{fig:sweep-results-all-neurons} and the comparison to LSTM in Table~\ref{tab:EGRU-LSTM-comparision}.

Beyond achieving strong decoding performance, the approach was designed for efficient hardware integration. Embedding the classifier within the recording frontend significantly reduces the volume of transmitted data. It enables real-time, low-latency closed-loop operation, which is essential for implantable systems facing power and bandwidth constraints~\cite{sussillo2012RNN_RL}. By processing signals close to the source, the system improves robustness against transmission artefacts and supports more autonomous and adaptive neuro-prosthetic applications. Overall, this work demonstrates that event-based recurrent architectures can combine high accuracy with hardware efficiency, paving the way for fully integrated neural interfaces.

\section{Conclusion}
We present an event-driven, recurrent neural decoder built on EGRU that achieves high task performance while maintaining a minimal resource footprint.
Through open-loop pretraining, efficient RL, and perturbation-aware training, our model demonstrates robust, real-time performance suitable for deployment in implantable neuroprosthetic systems.
This work advances the feasibility of on-device, closed-loop neural decoding, balancing the competing demands of accuracy, latency, and power efficiency. 
Future work will focus on hardware co-design and adaptation to human clinical datasets.

\section*{Acknowledgment}
The authors gratefully acknowledge the computing time made available to them on the high-performance computer at the NHR Center of TU Dresden. This center is jointly supported by the Federal Ministry of Education and Research and the state governments participating in the NHR (www.nhr-verein.de/unsere-partner).

\bibliographystyle{IEEEtran}
\bibliography{bibliography.bib}

\begin{thebibliography}{1}
\providecommand{\url}[1]{#1}
\csname url@samestyle\endcsname
\providecommand{\newblock}{\relax}
\providecommand{\bibinfo}[2]{#2}
\providecommand{\BIBentrySTDinterwordspacing}{\spaceskip=0pt\relax}
\providecommand{\BIBentryALTinterwordstretchfactor}{4}
\providecommand{\BIBentryALTinterwordspacing}{\spaceskip=\fontdimen2\font plus
\BIBentryALTinterwordstretchfactor\fontdimen3\font minus
  \fontdimen4\font\relax}
\providecommand{\BIBforeignlanguage}[2]{{%
\expandafter\ifx\csname l@#1\endcsname\relax
\typeout{** WARNING: IEEEtran.bst: No hyphenation pattern has been}%
\typeout{** loaded for the language `#1'. Using the pattern for}%
\typeout{** the default language instead.}%
\else
\language=\csname l@#1\endcsname
\fi
#2}}
\providecommand{\BIBdecl}{\relax}
\BIBdecl

\bibitem{Yik2025Neurobench}
``The neurobench framework for benchmarking neuromorphic computing algorithms
  and systems,'' \emph{Nature Communications}, vol.~16, no.~1, p. 1545, 2025.

\bibitem{subramoneyEGRU}
A.~Subramoney, K.~K. Nazeer, M.~Sch{\"o}ne, C.~Mayr, and D.~Kappel, ``Efficient
  recurrent architectures through activity sparsity and sparse back-propagation
  through time,'' in \emph{The Eleventh International Conference on Learning
  Representations}, 2023.

\bibitem{khan2024EGRUonS2}
K.~K. Nazeer, M.~Schöne, R.~Mukherji, B.~Vogginger, C.~Mayr, D.~Kappel, and
  A.~Subramoney, ``Language modeling on a spinnaker2 neuromorphic chip,'' in
  \emph{2024 IEEE 6th International Conference on AI Circuits and Systems
  (AICAS)}, 2024, pp. 492--496.

\bibitem{bakker2001RNN_RL}
B.~Bakker, ``Reinforcement learning with long short-term memory,''
  \emph{Advances in neural information processing systems}, vol.~14, 2001.

\bibitem{ni2022RNN_RL}
T.~Ni, B.~Eysenbach, and R.~Salakhutdinov, ``Recurrent model-free {RL} can be a
  strong baseline for many {POMDP}s,'' in \emph{International Conference on
  Machine Learning}.\hskip 1em plus 0.5em minus 0.4em\relax PMLR, 2022, pp.
  16\,691--16\,723.

\bibitem{sussillo2012RNN_RL}
D.~Sussillo, P.~Nuyujukian, J.~M. Fan, J.~C. Kao, S.~D. Stavisky, S.~Ryu, and
  K.~Shenoy, ``A recurrent neural network for closed-loop intracortical
  brain--machine interface decoders,'' \emph{Journal of neural engineering},
  vol.~9, no.~2, p. 026027, 2012.

\end{thebibliography}

\end{document}